\documentclass{ecai} 

\usepackage{latexsym}
\usepackage{amssymb}
\usepackage{amsmath}
\usepackage{amsthm}
\usepackage{booktabs}
\usepackage{enumitem}
\usepackage{graphicx}
\usepackage{color}
\usepackage{hyperref}

\newcommand{\BibTeX}{B\kern-.05em{\sc i\kern-.025em b}\kern-.08em\TeX}

\begin{document}


\begin{frontmatter}



\title{Symbolic Disentangled Representations for Images}


\author[A]{\fnms{Alexandr}~\snm{Korchemnyi}}
\author[A,B,C]{\fnms{Alexey K.}~\snm{Kovalev}\thanks{Corresponding Author. Email: kovalev@airi.net}}
\author[A,B,C]{\fnms{Aleksandr I.}~\snm{Panov}} 

\address[A]{Moscow Institute of Physics and Technology, Dolgoprudny, Russia}
\address[B]{AIRI, Moscow, Russia}
\address[C]{Federal Research Center “Computer Science and Control” of the Russian Academy of Sciences, Moscow, Russia}


\begin{abstract}
The idea of disentangled representations is to reduce the data to a set of generative factors that produce it. Typically, such representations are vectors in latent space, where each coordinate corresponds to one of the generative factors. The object can then be modified by changing the value of a particular coordinate, but it is necessary to determine which coordinate corresponds to the desired generative factor -- a difficult task if the vector representation has a high dimension.
In this article, we propose ArSyD (Architecture for Symbolic Disentanglement), which represents each generative factor as a vector of the same dimension as the resulting representation. In ArSyD, the object representation is obtained as a superposition of the generative factor vector representations. We call such a representation a \textit{symbolic disentangled representation}. We use the principles of Hyperdimensional Computing (also known as Vector Symbolic Architectures), where symbols are represented as hypervectors, allowing vector operations on them. Disentanglement is achieved by construction, no additional assumptions about the underlying distributions are made during training, and the model is only trained to reconstruct images in a weakly supervised manner.
We study ArSyD on the dSprites and CLEVR datasets and provide a comprehensive analysis of the learned symbolic disentangled representations. We also propose new disentanglement metrics that allow comparison of methods using latent representations of different dimensions.
ArSyD allows to edit the object properties in a controlled and interpretable way, and the dimensionality of the object property representation coincides with the dimensionality of the object representation itself.
\end{abstract}

\end{frontmatter}


\section{Introduction}
Good data representation for machine learning algorithms is one of the key success factors for modern approaches. Initially, the construction of good representations consisted of feature engineering, i.e., the manual selection, creation, and generation of such features that allow the model to successfully solve the main problem. Although feature engineering is still used in some areas, current models rely on learning representations from data~\cite{6472238}. This approach has become ubiquitous in computer vision~\cite{NIPS2012_c399862d}, natural language processing~\cite{DBLP:journals/corr/abs-1301-3781}, processing of complex structures such as graphs~\cite{Henaff2015DeepCN,2018graph}, and others.  A good representation can mean a representation with different properties, e.g. proximity of representations for semantically related objects~\cite{DBLP:journals/corr/abs-1301-3781}, identification of common features in objects~\cite{NIPS2012_c399862d}, preservation of a complex structure with dimensionality reduction, and disentanglement of representations~\cite{higgins2017betavae,NIPS2016_7c9d0b1f,pmlr-v80-kim18b}.

\begin{figure}[t]
    \centering
    \includegraphics[width=\linewidth]{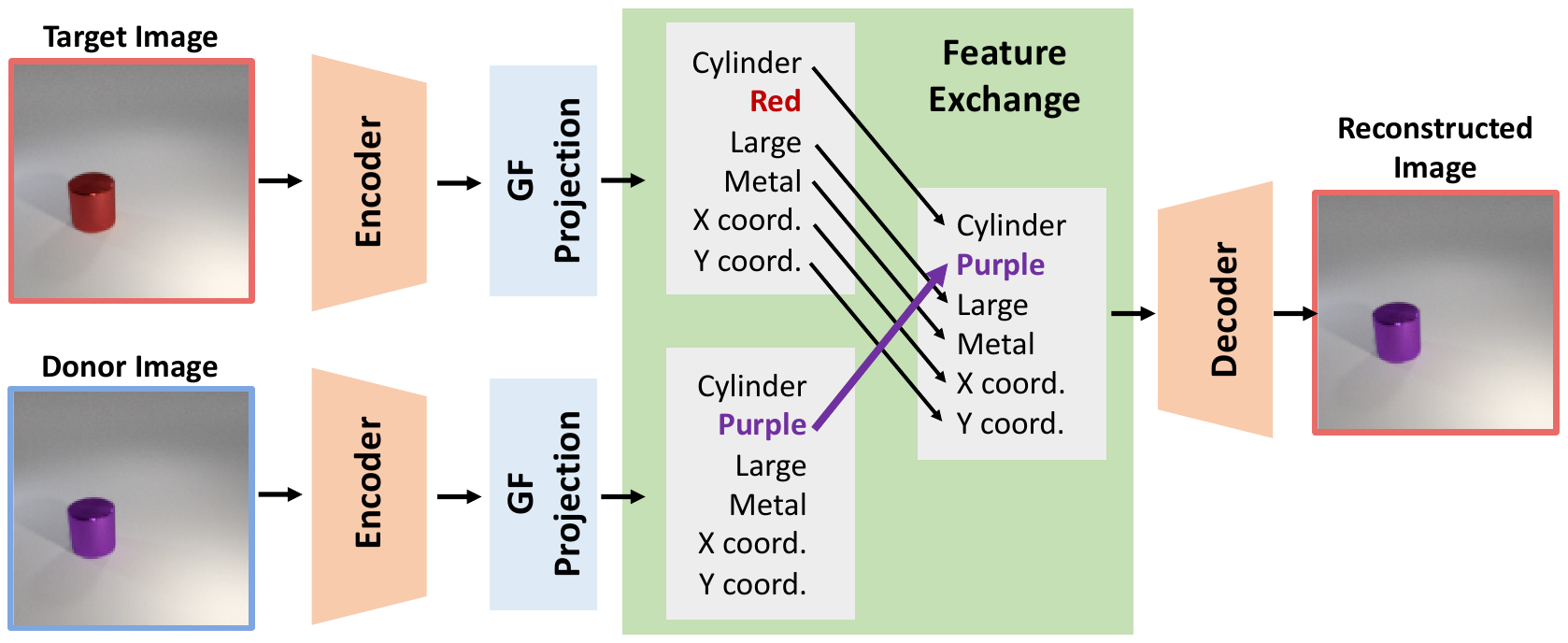}
    \caption{The Architecture for Symbolic Disentanglement (ArSyD) uses an encoder network and a Generative Factor (GF) projection to represent the target and donor images as sets of HVs. These HVs correspond to the values of the underlying generative factors in the data. The same encoder is used for both target and donor objects. While most of the values for the generative factors are the same between the donor and target images, one value is different. The Feature Exchange module exchanges the vector from the set of HVs for the target image associated with this different value for a vector from the set of HVs for the donor image. The decoder then reconstructs the target image using the desired value of the generative factor.}
    \label{fig:exchange_principle}
    \vspace{15pt}
\end{figure}

In this article, we consider disentangled representations. Following the work in~\cite{DBLP:conf/iclr/EastwoodW18}, we define disentangled representations as those described by the following three properties: 1) \textit{disentanglement}, i.e., the representation should factorize (disentangle) the underlying generative factors, so that one variable captures at most one factor, 2) \textit{completeness}, the degree to which each underlying factor is captured by a single variable, 3) \textit{informativeness}, i.e., the amount of information that a representation captures about the underlying generative factors (to be useful for tasks that require knowledge of the important data attributes, representations must fully capture information about the underlying generative factors). The disentangled representation can potentially improve generalization and explainability in many machine learning tasks: structured scene representation and scene generation~\cite{ElNouby2019TellDA,Matsumori2021UnifiedQT}, reinforcement learning~\cite{Keramati2018FastEW,kulkarni2019unsupervised,Berner2019Dota2W}, planning~\cite{Migimatsu2020ObjectCentricTA}, reasoning~\cite{NEURIPS2019_bc3c4a63,webb2023systematic}, and object-centric visual tasks~\cite{Groth2018ShapeStacksLV,Yi2020CLEVRERCE,Singh2021IlliterateDL,kirilenko2023quantized,kirilenko2024objectcentric}. The main research in disentanglement focuses on obtaining representations expressed by a vector, where each coordinate of the vector captures a generative factor~\cite{NIPS2016_7c9d0b1f,higgins2017betavae,pmlr-v80-kim18b}, and by changing this coordinate, we can change the property of the represented object.

In this work, we propose Architecture for Symbolic Disentanglement (ArSyD), which uses a disentangled representation in which an entire vector of the same dimension as the resulting representation of the object captures a generative factor. The object representation is then obtained as a superposition of the vectors responsible for each of the generative factors. We call such a representation a \textit{symbolic disentangled representation}.
ArSyD is based on the principles of Hyperdimensional Computing (HDC) (also known as Vector Symbolic Architectures)~\cite{Kanerva2009,kleyko2023surveyA,kleyko2023surveyB}, where symbols are represented by vectors of high and fixed dimension -- \textit{hypervectors} (HVs also known as \textit{semantic pointers}~\cite{Blouw2016ConceptsAS}). In most applications~\cite{10.1007/978-3-030-37487-7_6,DBLP:journals/corr/abs-2003-05171,kovalev2021applying,kirilenko2021question,KOVALEV202252,kirilenko2022vector}, to represent an object as a vector in high-dimensional latent space (the dimension $d$ is typically greater than 1000), HVs are randomly sampled from this space -- these vectors are called \textit{seed vectors}. However, some approaches learn vectors from data~\cite{Ozgur15,DBLP:journals/corr/abs-2110-08343}.

In HDC, as mentioned earlier, the seed vectors that form the description of an object and characterize its features are usually obtained by sampling from a predetermined high-dimensional space~\cite{10.1007/978-3-030-37487-7_6,KOVALEV202252}. They are fixed and are not learned from data. HDC takes advantage of the fact that with an extremely high probability, all seed vectors from high-dimensional spaces are dissimilar to each other (quasi-orthogonal), this is due to the concentration of measure phenomenon~\cite{ledoux2001concentration}. This allows us to reduce the manipulation of the symbols to vector operations. Two main operations in HDC are bundling (addition) and binding (multiplication). The bundling maps a set of HVs to the resulting vector, which is similar to summand vectors but quasi-orthogonal to other seed vectors. Semantically, the bundling represents a set of vectors and, thus, a set of symbols. The binding maps the two HVs to another vector that is dissimilar (quasi-orthogonal) to the summands and other HVs of the vector space. The binding can be interpreted as a representation of an attribute-value pair, i.e., the assignment of a value to a corresponding attribute. Thus, although the resulting representations are expressed as vectors, they also have symbolic properties. HDC uses a distributed representation, where the vector as a whole encodes an object or its property, rather than a separate coordinate of the vector as in a localist representation.

In ArSyD, the vectors representing the feature value of an object are obtained by applying the attention mechanism~\cite{bahdanau2014neural} over a set of fixed seed vectors (a codebook or an item memory in terms of VSA) and the initial representation of the object obtained using the encoder network. The training sample is a pair of objects that match in all but one property. Each object is split into vectors equal to the number of generative factors by an encoder and a generative factor projection network (GF Projection). Next, the feature vectors that differ between the target object and the donor object are exchanged. Then, the target object is reconstructed with a feature vector from the donor object (Figure \ref{fig:exchange_principle}). This weak supervision procedure~\cite{shu2019weakly,locatello2020weakly} allows the model to learn disentangled representations and edit objects in a controlled manner by manipulating their latent representations for cases where only one object is represented in the scene. For the scenes with multiple objects, we have combined the proposed approach with a Slot Attention model~\cite{locatello2020object}, which allows us to obtain object masks. These masks are used to extract a specific object from the scene. After that, we can edit the properties of a concrete object and restore the scenes already taking into account these changes.

Since ArSyD is based on the principles of HDC and uses a distributed representation, it is impossible to use popular disentanglement metrics (such as BetaVAE score~\cite{higgins2017betavae}, DCI disentanglement~\cite{DBLP:conf/iclr/EastwoodW18}, and others) for localist representations. Therefore, we propose two disentanglement metrics based on the classification of a reconstructed scene -- the Disentanglement Modularity Metric (DMM) and the Disentanglement Compactness Metric (DCM) -- that allow the evaluation of models regardless of the type of latent representation used. The DMM is responsible for the relationship between the latent unit (coordinate, vector, or other representation) and the generative factor. The DCM is reduced when a change in one unit in the latent space changes the classification of only one feature.

We have demonstrated how ArSyD learns symbolic disentangled representations for objects from modified dSprites~\cite{dsprites17} and CLEVR datasets~\cite{johnson2017clevr}, and provide the proof-of-concept results on the realistic CelebA dataset~\cite{liu2015faceattributes}. We also used ArSyD combined with the Slot Attention model to edit scenes from the CLEVR datasets.

Thus, the contributions of this article are:
\begin{enumerate}
    \item An Architecture for Symbolic Disentanglement (ArSyD), which uses a weakly supervised approach to learn symbolic disentangled representations based on a combination of attention over a codebook of fixed vectors and the principles of HDC;
    \item The learned representations allow us to edit objects in a controlled and interpretable way by manipulating their representations in the latent space;
    \item A new model based on Slot Attention that allows object-centric and interpretable editing of a scene with multiple objects;
    \item New disentanglement metrics that allow the comparison of models with different types of latent representations.
\end{enumerate}

\section{Disentangled Representation}
Most of the methods for learning disentangled representations are based on either the Variational Autoencoder (VAE)~\cite{higgins2017betavae,https://doi.org/10.48550/arxiv.1804.03599,pmlr-v80-kim18b,NEURIPS2018_1ee3dfcd,DBLP:conf/iclr/0001SB18} or on the Adversarial Generative Networks~\cite{https://doi.org/10.48550/arxiv.1406.2661} (GAN)~\cite{NIPS2016_7c9d0b1f,DBLP:journals/corr/abs-1906-06034,DBLP:journals/corr/abs-2003-03461}.

In these approaches, the disentanglement is achieved by imposing additional constraints on the loss: by introducing a $\beta$ parameter to balance the independence constraints with reconstruction accuracy in BetaVAE~\cite{higgins2017betavae}; by adding a capacity control objective in~\cite{https://doi.org/10.48550/arxiv.1804.03599}; by factorizing the representation distribution in FactorVAE~\cite{pmlr-v80-kim18b}; by decomposing the evidence lower bound in $\beta$-TCVAE~\cite{NEURIPS2018_1ee3dfcd}; by minimizing the covariance between the latents in DIP-VAE~\cite{DBLP:conf/iclr/0001SB18}; by maximizing the mutual information in InfoGAN~\cite{NIPS2016_7c9d0b1f}; by introducing a contrastive regularizer in InfoGAN-CR~\cite{DBLP:journals/corr/abs-1906-06034}; by adding a mutual information loss to StyleGAN~\cite{DBLP:journals/corr/abs-1812-04948} in Info-StyleGAN~\cite{DBLP:journals/corr/abs-2003-03461}. Other approaches impose additional restrictions based on group theory on existing VAE models~\cite{yang2022towards}, or propagate inductive regulatory bias recursively over the compositional feature space~\cite{chen2022recursive}, or provide a framework for learning disentangled representations and discovering the latent space~\cite{ren2022learning}.

In ArSyD, we do not impose additional specific restrictions on the loss, but use a structured representation of the object in the latent space using HDC principles and a special learning procedure. We represent each generative factor as a HV of the same dimension as the resulting representation. The object representation is then obtained as a superposition of the vectors responsible for the generative factors.

To evaluate the disentanglement of the representation, special metrics are used that assume a localist representation in the latent space, where each coordinate represents the generative factor. The BetaVae score~\cite{higgins2017betavae} uses a simple classifier to predict the index of a generative factor that is held fixed while other factors are randomly sampled. This approach is not applicable to distributed representations because individual positions in such representations do not carry all the information about the generative factor. FactorVAE score~\cite{pmlr-v80-kim18b} proposes a similar approach, but uses an index of a position with less variance.  MIG~\cite{NEURIPS2018_1ee3dfcd} uses mutual information and also refers to the individual coordinates of the latent representation, making it impossible to use this metric for other representations. SAP score~\cite{DBLP:conf/iclr/0001SB18} uses classification or linear regression, which also does not work for distributed representations.
In this article, we propose two new disentanglement metrics that allow the comparison of models regardless of whether they use a localist or a distributed representation in the latent space.

\section{Methods}
\subsection{Object Representation in Latent Space}
\label{subsec:objrepr}
In this article, we apply HDC principles to obtain an object representation in the latent space and then map this representation to the raw data. We use a distributed representation and represent each generative factor as an HV with the same dimensionality as a final object representation. This means that the information is distributed across all components, no single HV component has a meaning, i.e., only the whole HV can be interpreted. This differs from the localist representation used by modern disentanglement models, where a single vector component potentially has a meaning. The nature of HVs can be different, such as binary~\cite{Kanerva2009}, real~\cite{Gayler,HRR_Plate,PlateBook,Eliasmith13,Komer19}, complex~\cite{Gayler}, or bipolar~\cite{Gayler2009ADB,Kleyko2018}. Also, the exact implementation of the vector operations may vary for different vector spaces while maintaining the computational properties.  For a detailed comparison of different VSA implementations, we recommend to refer to~\cite{schlegel2022comparison}.

To represent a simple entity in HDC, the seed vector is sampled from the vector space and stored in an item memory. When an object has a complex structure, the resulting representation is obtained by performing vector operations on the seed vectors from the item memory~\cite{kleyko2023surveyA}. We illustrate the vector operations defined in HDC with an example from the Holographic Reduced Representation (HRR)~\cite{HRR_Plate} that we use in this article. This implementation works with real-valued HVs sampled from the normal distribution $\mathcal{N}\left(0, \frac{1}{D}\right)^D$ with mean $0$ and variance $\frac{1}{D}$, and uses cosine distance as the similarity measure, where $D$ is the vector dimension.

Two main vector operations are bundling (addition) and binding (multiplication). The \textit{addition} ($+$) is implemented as an element-wise sum with normalization: $R = \frac{\sum_{i=1}^{N}V_i}{N}$. The resulting vector $R$ is similar to the summand vectors $V_i$, but quasi-orthogonal to other seed vectors. Semantically, bundling represents a set of vectors and, thus, a set of symbols. The \textit{multiplication} ($\circledast$) uses a circular convolution: $R = V_1 \circledast V_2 = \mathcal{F}^{-1}\{\mathcal{F}\{V_1\} \odot \mathcal{F}\{V_2\}\}$, where $\odot$ is an element-wise multiplication, $\mathcal{F}$ and $\mathcal{F}^{-1}$ are the Fourier transform and the inverse Fourier transform, respectively. The multiplication maps the vectors $V_1$ and $V_2$ to another HV -- $R$. The resulting vector $R$ is dissimilar (quasi-orthogonal) to $V_1$, $V_2$ and other HVs of the vector space. Binding represents an attribute-value pair, i.e. the assignment of a value to a corresponding attribute.

\begin{figure*}[ht]
    \centering
    \includegraphics[width=\linewidth]{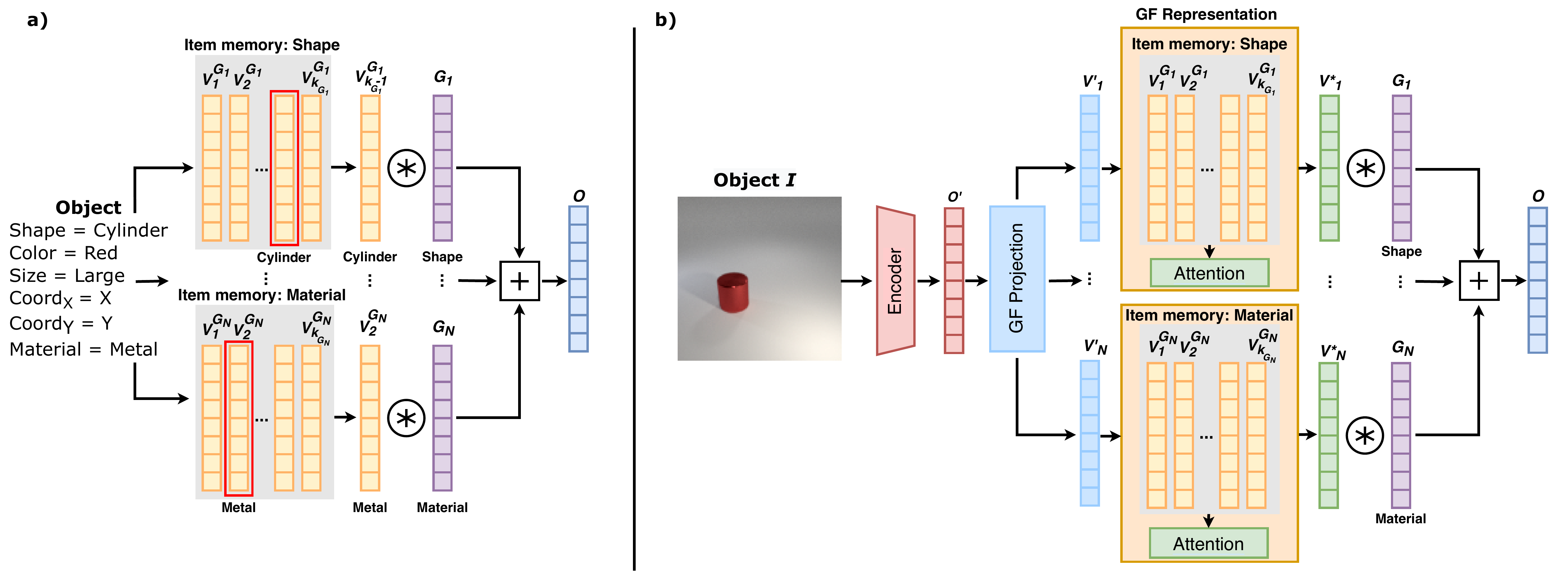}
    \caption{The process of obtaining a symbolic disentangled representation of an object: \textit{\textbf{a)}} usual approach in HDC -- a symbolic description is used without grounding in the raw data; \textit{\textbf{b)}}~approach in ArSyD -- the symbolic description of an object is grounded in the raw data to obtain its symbolic disentangled representation.}
    \label{fig:obj_representation}
    \vspace{10pt}
\end{figure*}

Using these operations, we can represent any object $O$ generated by $N$ underlying generative factors as a set of attribute-value pairs (Figure~\ref{fig:obj_representation}a):
\begin{equation}
\begin{gathered}
\label{eq:set_representation}
    O = G_1 \circledast V_{2}^{G_1} + G_2 \circledast V_{4}^{G_2} + \dots + G_N \circledast V_{1}^{G_N},
\end{gathered}
\end{equation}
\noindent where $G_i$ is a $i$-th generative factor, $V_{j_m}^{G_i}$ is a $j_m$-th value of a generative factor $G_i$. $j_m = 1 \dots k_{G_i}$, where $k_{G_i}$ is the number of values of a generative factor $G_i$.

Thus, the target object $O$ in Figure~\ref{fig:exchange_principle} can be represented as:
\begin{equation}
\begin{gathered}
O = S \circledast Cy + C \circledast R + Si \circledast L + M \circledast Me + \\ + C_X \circledast X + C_Y \circledast Y,
\end{gathered}
\end{equation}
\noindent where $(S)hape, (C)olor, (Si)ze, (M)aterial, (C)oord_X,\\  (C)oord_Y$ and $(Cy)linder, (R)ed, (L)arge, (Me)tal, X, Y$ are HVs for underlying generative factors and their corresponding values. The HVs for the generative factors $G_i$ and the values $V_{k_{G_i}}^{G_i}$ are sampled at the time of model initialization, then fixed and stored in the corresponding item memory (codebook). They do not change during the training and testing phases of the model. The number of sampled HVs $G_i$ corresponds to the number of generative factors, while the number of HVs $V_{k_{G_i}}^{G_i}$ corresponds to the number of values for the respective generative factor $G_i$. Thus, we can obtain an HDC representation of an object according to its symbolic description without grounding in the raw data. This approach has proven effective in many applications, but it requires an explicit algorithm to represent an object by its properties, which makes it difficult to transfer to new domains.

In this article, we propose an architecture that allows HDC representations to be grounded in the raw data (Figure~\ref{fig:obj_representation}b). The first step is to use an encoder to extract an intermediate object representation $O'$ from the image $I$. Next, we use the Generative Factor (GF) projection module to obtain an intermediate representation of the generative factor values $V'_i$. We use a fully connected neural network with $D$ inputs and $ND$ outputs, where $N$ is the number of underlying generative factors for this module. The output of this module is decomposed into $N$ vectors, which are $V'_1, \dots, V'_N$ vectors of dimension $D$. These vectors are then fed into the Generative Factor representation modules, which use the attention mechanism \cite{bahdanau2014neural} to represent the value vector $V^{*}_i$ as a linear combination of seed vectors from the codebook $V_{j_m}^{G_i}$:
\begin{equation}
\label{eq:att_1}
\begin{gathered}
    a_\ell = softmax\left(\frac{V'_i K_{j_m}}{\sqrt{D}}\right)_{j_m},
\end{gathered}
\end{equation}
\noindent where $softmax()_{j_m}$ -- the ${j_m}$-th component, $D$ -- the dimension of a vector space, $K_{j_m}$ -- the projection of the ${j_m}$-th seed vector $V_{{j_m}}^{G_i}$ from the item memory,
\begin{equation}
\label{eq:att}
\begin{gathered}
    V^{*}_i = a_1 V_{1}^{G_i} + a_2 V_{2}^{G_i} + \dots + a_{k_{G_i}} V_{k_{G_i}}^{G_i}.
\end{gathered}
\end{equation}
The main challenge is to bridge the gap between localist and distributed representations. This is necessary to effectively ground distributed HDC representations in the raw data. In ArSyD, this is achieved by using the attention mechanism (Equation~\ref{eq:att_1}). Frozen vectors $V^{G_i}_{k_{G_i}}$ stored in the item memory and initially sampled with respect to the distributed representation are used as value vectors. The trained projections $V'_i$ and $K_{j_m}$ make it possible to build a bridge between the localist representations $O'$, obtained from the encoder output, and the distributed representations of $V^{G_i}_{k_{G_i}}$, stored in the item memory, without changing the latter. Additional training of the $V^{G_i}_{k_{G_i}}$ vectors leads to a localist representation that violates the original premises.

The resulting value vectors $V^{*}_i$ are multiplied by the corresponding generative factor vectors $G_i$ and summed to produce the vector $O$, which is a symbolic disentangled representation of an object. The resulting vector $O$ is used to decode the object into an image. The Encoder, GF Projection, and Attentions modules are trainable modules.

\subsection{Weakly supervised learning}
We use a weakly supervised learning approach for our model, as done by other works~\cite{ bouchacourt2018multi,jha2018disentangling,esser2019unsupervised,sanchez2020learning,vowels2020nestedvae}, which is shown \cite{shu2019weakly,locatello2020weakly} to allow us to learn disentangled representations. It is assumed that the object $O$ depicted on the image $I$ can be represented as a set of $N$ generative factors $G_i$ with a corresponding values $V_{j_m}^{G_i}$ (Equation~\ref{eq:set_representation}).

If we take two objects $O_1$ and $O_2$ and encode them into a set of generative factors $G^1=G(O_1)$ and $G^2=G(O_2)$, we want that the vectors $V_{j_m}^{G^1_i} = V_{j_m}^{G^2_i}$ if $G^1_i = G^2_i$ for all indices $i$ but one, e.g., index $p$. Then we can exchange the values at index $p$ between the target object and the donor object and get the representations $\hat{O_1}$ and $\hat{O_2}$, respectively.

If we reconstruct the original object $O_1$ from the representation $\hat{O_1}$, the result of the reconstruction should not differ from the original donor object $O_2$. Similarly, if we reconstruct the object from the representation $\hat{O_2}$, we get the object $O_1$. In this way, we can build a learning model. Since each training example has objects $O_1$ and $O_2$, the reconstruction possibility is symmetric for both objects, and we can use two reconstructions at once: $\hat{O_1}$ compared to $O_2$, and $\hat{O_2}$ compared to $O_1$.

\subsection{End-to-end Model}
For experiments with a single object in the scene, we use the architecture shown in Figure~\ref{fig:obj_representation}b and add a decoder that reconstructs the images of the target and donor objects. We use only the reconstruction losses, the mean squared error (MSE) between the original and reconstructed images, to train the model:
\begin{equation}
\begin{gathered}
\label{eq:single_object_loss}
L_{Object} = MSE(O_t, \tilde{O_t}) + MSE(O_d, \tilde{O_d}),
\end{gathered}
\end{equation}
\noindent where $O_t$ -- a target object, $\tilde{O}_d$ -- a reconstructed target object, $O_d$ -- a donor object, $\tilde{O}_d$ -- a reconstructed donor object.

\begin{figure}[t]
    \centering
    \includegraphics[width=\linewidth]{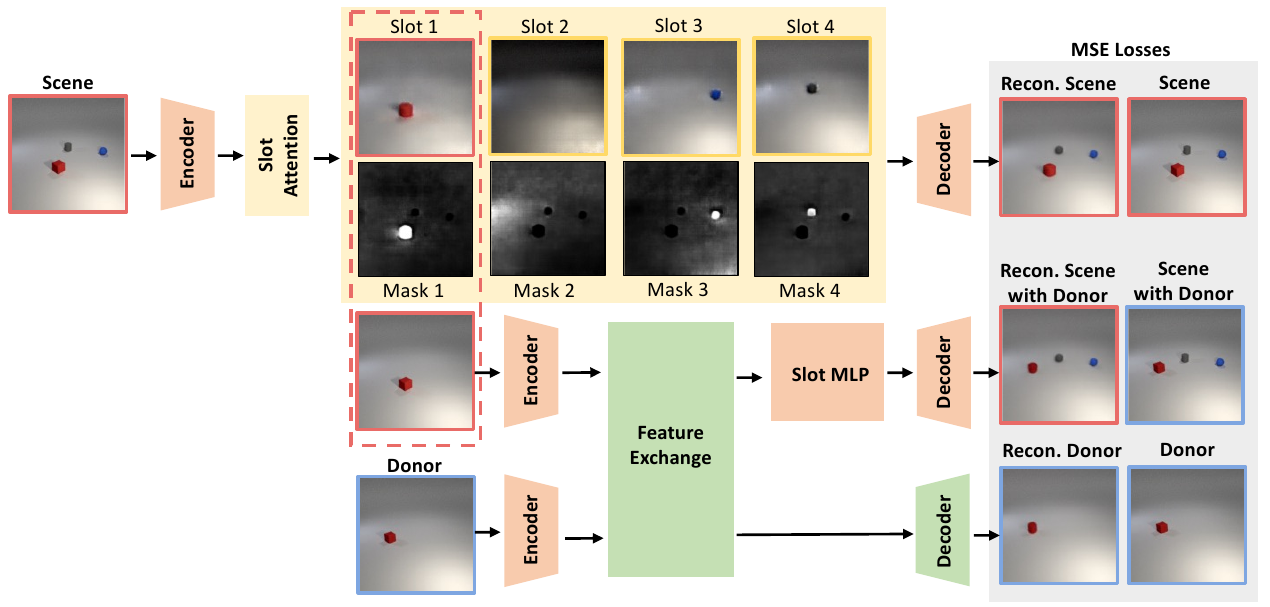}
    \caption{An ArSyD architecture for the case of multiple objects in the scene. We use Slot Attention to discover objects in the scene and represent it as a collection of individual objects. The Feature Exchange module is then used to replace the value of the desired generative factor. The model is trained end-to-end with a loss $L_{Scene}$ (Equation~\ref{eq:scene_loss}).}
    \label{fig:fig3_multiple_objects}
    \vspace{20pt}
\end{figure}

The overall end-to-end model for the case of multiple objects in the scene is shown in Figure \ref{fig:fig3_multiple_objects}. We use Slot Attention~\cite{locatello2020object} to discover objects in the scene and represent it as a collection of individual objects. Then, for the target object, we reconstruct its image from a corresponding slot $S$ and use this image in the feature exchange module. The Feature Exchange module is implemented in the same way as in Figure \ref{fig:exchange_principle}, except that that instead of reconstructing the obtained object representation with a changed value of the generative factor, we use a Multilayer Perceptron (MLP) to map it into the slot space -- $S'$. Then, we replace the original slot of the object $S$ with its modified version $S'$ and reconstruct the scene image. The reconstructed scene contains an object with a modified value of the generative factor.

During training, we only use reconstruction losses (MSE) to control the quality of the image restoration:
\begin{equation}
\begin{gathered}
\label{eq:scene_loss}
L_{Scene} = MSE(S, \tilde{S}) + MSE(S', \tilde{S'}) + MSE(O_d, \tilde{O_d}),
\end{gathered}
\end{equation}
\noindent where $S$ -- an original scene, $\tilde{S}$ -- a reconstructed original scene, $S'$ -- a scene with donor, $\tilde{S'}$ -- a reconstructed scene with donor, $O_d$ -- a donor object, $\tilde{O}_d$ -- a reconstructed donor object.

We aimed to minimize the need for auxiliary information, thus we use only original images and their reconstructions. The first term, $MSE(S, \tilde{S})$, facilitates the training of the Slot Attention module. The second term, $MSE(S', \tilde{S'})$, is crucial for linking both the Slot Attention and the Feature Exchange modules together with the Slot Attention MLP module. The third term, $MSE(O_d, \tilde{O_d})$, is used to train the Feature Exchange module. 

\subsection{Evaluation Metrics}
Popular disentanglement metrics such as BetaVAE score~\cite{higgins2017betavae}, DCI disentanglement~\cite{DBLP:conf/iclr/EastwoodW18}, MIG~\cite{NEURIPS2018_1ee3dfcd}, SAP score~\cite{DBLP:conf/iclr/0001SB18}, and FactorVAE~score \cite{pmlr-v80-kim18b} are based on the assumption that disentanglement is achieved by each individual vector coordinate capturing a generative factor in the data -- localist representations. Since we use distributed representations in ArSyD, current metrics are not suitable for evaluating the disentanglement of the proposed model representations. When comparing models with different latent space representation structures (e.g., localist or distributed representations), the problem arises that a direct comparison using metrics based on the latent representations themselves is not possible. For this reason, we propose new metrics that allow to evaluate different models for disentangled representations, regardless of how the individual generative factor is represented. In the proposed approach, the metrics are measured using an explicit representation rather than a latent representation, e.g., in pixel space. Although we use images in our work, this approach can be extended to other modalities (text, sound, and others).

We use the term \textit{unit of the latent representation} (or \textit{unit} for short) to refer both to the value of the coordinate of the latent representation vector for the localist representation, and to an HV selected from the codebook that is responsible for a particular generative factor for the distributed representation.

The main idea is that if we have a latent disentangled representation of an object, we can control the representation of that object in pixel space. In other words, by changing the values of the unit responsible for the color of the object, the color and only the color of the reconstructed object should change. If the representation is not disentangled, other properties of the object may change. Also, if the latent unit is not responsible for any of the generative factors, none of the properties may change. Thus, we can define metrics based on the classification of the reconstructed images in pixel space, while changing individual generative factors in latent space.

Knowing the number and type of the generative factors, we can train models (classifiers) on the training set that will predict the values of the generative factors from object representations in pixel space. For the validation set, we first obtain a latent representation of the object $h$, reconstruct its representation in pixel space ($Decoder(h) = \hat{S}$), and predict the values of the generative factors from it -- $Cl(\hat{S}) = \hat{y}$. It is important to note that these predictions may not match the ground truth answers for the original image due to decoder errors. Next, we change the latent representation of the object in a controlled way (by changing the unit $v$), obtain $h'$, reconstruct the image ($Decoder(h') = \hat{S}'_{v}$), and obtain predictions for the values of the generative factors on it: $Cl(\hat{S}'_{v}) = \hat{y}'$. We then compare these values with previous predictions -- $[\hat{y}' \neq \hat{y}]$ -- and write 0 if the prediction for the corresponding factor has not changed, and 1 if it has. Note that we do not use the ground truth answers of the original images, which allows us to negate the effects of a possible poor quality of the decoder reconstruction (if the wrong shape was reconstructed both times during decoding, this will not be taken into account in the metrics).

\begin{figure}[ht]
    \centering
    \includegraphics[width=0.95\linewidth]{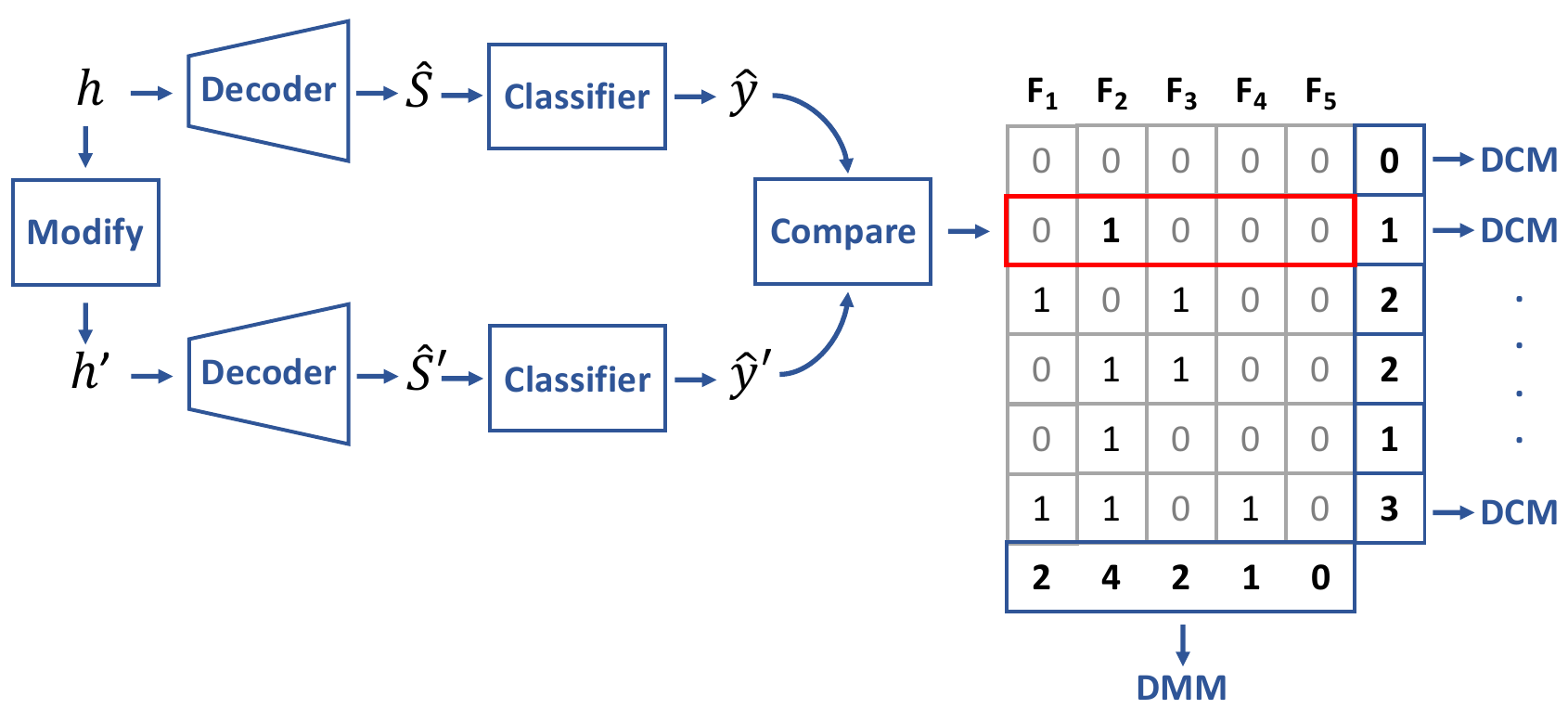}
    \caption{The principle of calculating the DMM and DCM metrics. We can represent the relations between generative factors and units in the latent space as a table: columns $F_1, \dots F_5$ correspond to the different generative factors, rows correspond to the different units $c$ of the latent representation. We write $1$ if $\hat{y}_i \neq \hat{y}'_i$ (the prediction for the corresponding factor has changed after changing the unit of the latent representation), otherwise it is $0$ (the prediction for the corresponding factor has not changed). We use the row sum to compute the DMM metric (when the same latent unit is changed many times) and the column sum to compute the DCN metric.}
    \label{fig:fig4_metrics}
    \vspace{20pt}
\end{figure}

Based on \cite{higgins2018definition} and the above mentioned ideas, we propose the Disentanglement Modularity Metric (DMM)~(Equation~\ref{eq:dmm}) and the Disentanglement Compactness Metric (DCM)~(Equation~\ref{eq:dcm}):
\begin{equation}
\begin{gathered}
\label{eq:dmm}
DMM = \frac{1}{|C|}\sum_{c \in C}{H(\sigma\sum_{v \in V_c}[Cl(\hat{S}'_v) \neq \hat{y}])}
\end{gathered}
\end{equation}

\begin{equation}
\begin{gathered}
\label{eq:dcm}
DCM = \frac{1}{|C|}\sum_{c \in C}|\sum_{G_i \in G}[Cl(\hat{S}'_c) \neq \hat{y})]_{G_i} - 1|,
\end{gathered}
\end{equation}
\noindent where $c$ -- a unit (coordinate, vector, or other representation) of the latent representation, $H(X)$ -- entropy, $\sigma$ -- softmax, $V_c$ -- set of all values to be checked for $c$, $Cl$ -- a classifier, $\hat{S}'_v$ -- a reconstruction of a latent representation for scene $S$ with modified latent value $v$, $\hat{y} = Cl(\hat{S})$ -- a predicted class for the reconstruction of the original scene $S$, and $[\;]$ -- an Iverson bracket. The subscript $i$ indicates that the classification of a particular generative factor $G_i \in G$ is considered. The principle of calculating the DMM and DCM metrics is shown in Figure~\ref{fig:fig4_metrics}.

The \textit{Disentanglement Modularity Metric} refers to evaluating whether each latent unit is responsible for encoding only one generative factor~\cite{higgins2018definition}. We select a unit $c$ from the latent representation and vary it with possible values in the latent space. After changing the latent vector, the original image is reconstructed and classified, and then compared to the classification result of the original reconstructed latent representation. Ideally, only one of the vector prediction coordinates would change during classification. Therefore, when normalizing and applying the softmax function, the vector of change in classifier values should consist of 1 in one coordinate and 0 in the others. When interpreted as a probability, this results in a minimum entropy. 

The \textit{Disentanglement Compactness Metric} measures whether each generative factor is encoded by a single latent unit~\cite{higgins2018definition}. This metric counts the number of classes that have changed after each change in the latent representation, and then normalizes the resulting value. Note that we do not normalize to the number of latent units (coordinates or vectors), which may motivate using fewer of them.

\label{sec:exp}
\subsection{Datasets}
In this article, we used datasets generated from dSprites~\cite{dsprites17} and CLEVR~\cite{johnson2017clevr} to test the proposed approach. To demonstrate the potential generalizability of our approach to non-synthetic real-world images, we used the CelebA dataset~\cite{liu2015faceattributes}.

\subsubsection{dSprites paired}
dSprites contains 737280 procedurally generated 2D shapes (images of size $64 \times 64$) with the following generative factors: shape (square, ellipse, heart), scale (6 values), orientation (40 values in $[0.2\pi]$), x and y coordinates (32 values for each).

Each training example contains two images $x_1, x_2$ with generative factors $(G^i_{1}, ... , G^i_{5})$ and a feature exchange vector $e = (e_1, ..., e_5)$, where $e_i = 1 - [G^1_{i}=G^2_{i}]$ ($[\;]$ is the Iverson bracket). Each of the two images can be a donor for the other (Figure~\ref{fig:exchange_principle}), and the feature exchange vector is used in the latent representation stage to exchange the corresponding features. 

For our training set, we sampled 100,000 pairs of unique images that differ by the value of one generative factor. At the same time, we excluded from the training set images with $shape = square$ and $x > 0.5$, to test ArSyD for compositional generalization as it was done in~\citet{lostinlatentspace2022}. The test set was sampled without restrictions and contained 30,000 pairs of unique images that differ by one value of the generative factors and do not match the examples from the training set. We call the resulting dataset \textbf{dSprites paired}. Examples of image pairs from the dSprites paired dataset are shown in Figure~\ref{fig:datasets}a.

\subsubsection{CLEVR paired}
CLEVR contains images of 3D-rendered objects with the following generative factors: shape, size, material type, color, x and y coordinates, and orientation (because the sphere and the cylinder are symmetric in the $XY$ plane, it is impossible to check the result when exchanging the "orientation" feature, therefore orientation was set to zero for all images and excluded from the list of generative factors).

\begin{figure}
    \centering
    \includegraphics[width=\linewidth]{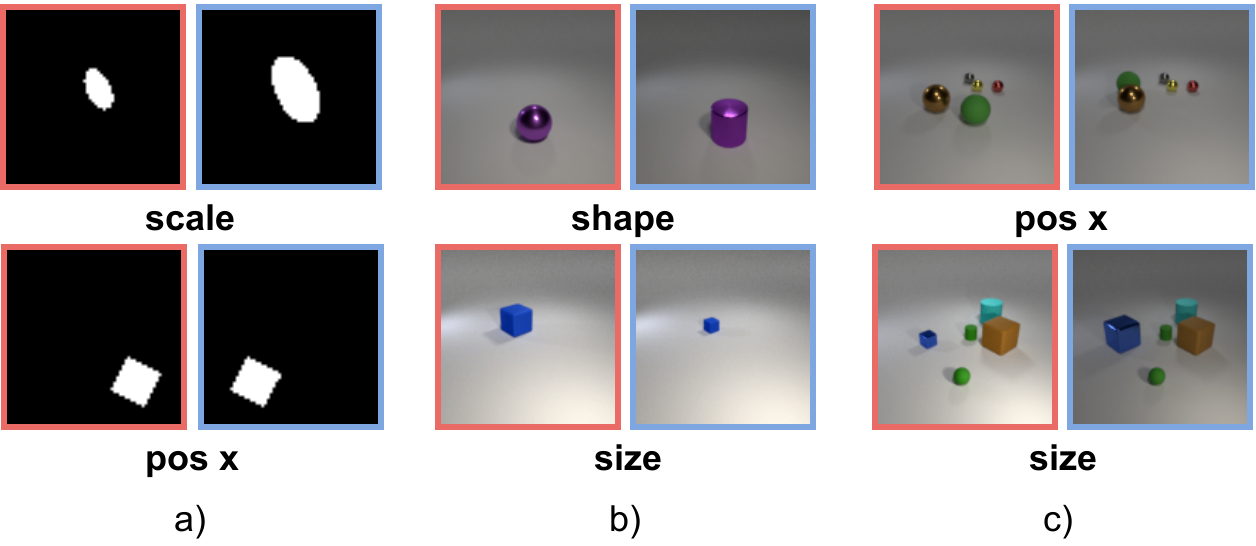}
    \caption{Examples of image pairs from the training sample of datasets: a) \textbf{dSprites paired}; b) \textbf{CLEVR1 paired}; c) \textbf{CLEVR5 paired}.}
    \label{fig:datasets}
    \vspace{10pt}
\end{figure}

We have modified the source generation code from the original dataset~\cite{johnson2017clevr} to create our training and test examples (10,000 and 1000, respectively). The first split, in which, as in dSprites, each training example contains two images $x_1, x_2$ and the feature exchange vector $e$ we call \textbf{CLEVR1 paired} (Figure~\ref{fig:datasets}b) because the images contain one object. The second split additionally contains random objects $(O_1, ... , O_k), k \in \{0, ..., 4\}$ and two scene images $S_1, S_2$ rendered from a set of objects, where $S_i = \{X_i, O_1, ... , O_k\}$, called \textbf{CLEVR5 paired} (Figure~\ref{fig:datasets}c). Camera position and lighting are random but fixed for all images of the same training example.

We used a version of CLEVR with color/shape conditions (CoGenT) to test for compositional generalization. The training set contains cubes that are gray, blue, brown, or yellow; cylinders that are red, green, purple, or cyan; and spheres that can be any color. In the validation set, the cubes and cylinders have opposite color palettes, and the test set contains all possible combinations. In addition, the size of the generated images is $128 \times 128$ (default is $320 \times 240$), and there is no restriction on overlapping objects.

\subsubsection{CelebA paired}
The CelebA dataset consists of 200k celebrity images and is labeled with 40 binary features describing appearance. Since ArSyD is trained in weakly supervised mode, we compiled the CelebA paired dataset from paired images, leaving only 7 features, 4 of which were obtained by grouping 11 features from the original label (``Hair'' = \{``Bald'', ``Black\_Hair'', ``Blond\_Hair'', ``Brown\_Hair'', ``Gray\_Hair''\}, ``Beard'' = \{``No\_Beard'', ``Goatee''\}, ``Nose'' = \{``Big\_Nose'', ``Pointy\_Nose''\}, ``Hair\_Form'' = \{``Straight\_Hair'', ``Wavy\_Hair''\}), and we use 3 features (``Male'', ``Young'', and ``Pale\_Skin'') as is. We kept only those images for which only one value for the grouped features is presented in the image. Thus, we obtained 28K images, of which we made 10k pairs (or 20k images), which is the same as the settings for CLEVR1. We call this version the \textbf{CelebA paired} dataset. We also allowed the images to differ in several generative factors. 

\subsection{Architecture Details}
In the experiments, we used simple convolutional neural network (CNN) encoder and decoder architectures presented in Table~\ref{table:cnn_encoder} and \ref{table:cnn_decoder}, respectively. For the K and Q projections in the attention mechanism, we used Linear layers of size 1024 without any nonlinearity.

\begin{table}[ht]
  \begin{center}
  \begin{tabular}{lccc}
    \toprule         
    \multicolumn{1}{c}{\textbf{Layer}}     & \textbf{Channels}    & \textbf{Activation} & \textbf{Params}  \\
    \midrule
    Conv2D 4 $\times$ 4  & 64  & ReLU & stride 2, pad 1 \\
    Conv2D 4 $\times$ 4  & 64  & ReLU & stride 2, pad 1 \\
    Conv2D 4 $\times$ 4  & 64  & ReLU & stride 2, pad 1 \\
    Conv2D 4 $\times$ 4  & 64  & ReLU & stride 2, pad 1 \\
    Conv2D 4 $\times$ 4 *  & 64  & ReLU & stride 2, pad 1 \\
    Linear  & 1024  & ReLU & -- \\
    Linear  & 1024  & -- & -- \\
    \bottomrule
  \end{tabular}
  \end{center}
  \caption{Architecture of the CNN encoder used in the experiments. *-- the layer is used in the models for CLEVR1 paired and CelebA paired.}
  \label{table:cnn_encoder}
\end{table}

\begin{table}[ht]
  \begin{center}
  \begin{tabular}{lccc}
    \toprule             
    \multicolumn{1}{c}{\textbf{Layer}}     & \textbf{Channels}    & \textbf{Activation} & \textbf{Params}  \\
    \midrule
    Linear  & 1024  & GELU & -- \\
    Linear  & 1024  & GELU & -- \\
    ConvTranspose2D 4 $\times$ 4  & 64  & GELU & stride 2, pad 1 \\
    ConvTranspose2D 4 $\times$ 4  & 64  & GELU & stride 2, pad 1 \\
    ConvTranspose2D 4 $\times$ 4  & 64  & GELU & stride 2, pad 1 \\
    ConvTranspose2D  4 $\times$ 4*  & 64  & GELU & stride 2, pad 1 \\
    ConvTranspose2D 4 $\times$ 4 & 64  & Sigmoid & stride 2, pad 1 \\
    \bottomrule
  \end{tabular}
  \end{center}
  \caption{Architecture of the CNN decoder used in the experiments. *-- the layer is used in the models for CLEVR1 paired and CelebA paired.}
  \label{table:cnn_decoder}
\end{table}

\subsection{Training}
We trained ArSyD using the AdamW optimizer~\cite{adamw} with a learning rate of $2.5e-5$ and other parameters set to default. We also use the learning rate scheduler OneCycleLR~\cite{onecyclelr} with the percentage of the cycle spent increasing the learning rate set to 0.2. For dSprites paired the number of epochs was 600 with a batch size of 512 and for CLEVR1 paired and CelebA paired, 1000 and 64 respectively.

For CLEVR5, the model is trained end-to-end and uses the Slot Attention module for the Object Discovery task. We trained a model over 2000 epochs, which is equivalent to 300k steps with a batch size of 64. For the Slot Attention module, we used parameters based on the parameters of the original article: number of iterations $T = 3$, and the number of slots $K = 6$, which is $1$ more than the maximum number of objects in the image.

It takes about 2.5 hours on a single NVIDIA TITAN RTX GPU to train our model on CLEVR1 for 600 epochs, 3.5 days on CLEVR5 for 2000, and about 2 hours on dSprites.

\subsection{Baselines and Metrics}
We compare ArSyD with BetaVAE~\cite{higgins2017betavae} and FactorVAE~\cite{pmlr-v80-kim18b} as prominent representatives of approaches to obtaining disentangled representations.

Both the DMM and DCM metrics require a pre-trained classifier, for which we used 6 separate pre-trained ResNet-34~\cite{he2016residual} models fine-tuned on the CLEVR1 paired dataset, each trained to classify values of a different generative factor.

We select a unit $c$ from the latent representation and vary it with possible values in the latent space. For ArSyD, this is a set of predefined vectors, for BetaVAE and FactorVAE models it is 32 linearly distributed values in the range of the $\mu \pm 3 \sigma$ for this latent coordinate. After changing the latent vector, the original image is reconstructed and classified, and then compared to the classification result of the original reconstructed latent vector. Ideally, only one of the coordinates would change during classification. Therefore, when normalizing and applying the softmax function, the vector of change in classifier values should consist of 1 in one coordinate and 0 in the others. This, when interpreted as a probability, gives a minimum entropy. 

We also provide quantitative results that evaluate the quality of the reconstruction of objects and scenes, along with the FID metric~\cite{NIPS2017_8a1d6947} for the CLEVR paired dataset and the Intersection over Union (IoU) for the dSprites paired dataset. For the preliminary experiments on the CelebA dataset, we provide qualitative results of image reconstruction.

\section{Results}
\label{sec:results}
In the series of experiments, we conducted in this article, we sought to answer the following research questions (RQ).

\paragraph{RQ1. Is it possible to obtain disentangled representations only with HDC?}
To test the hypothesis of obtaining disentangled representations using HDC, we trained only the decoder (without feature exchange) on CLEVR1 paired and dSprites paired representations obtained without resorting to the raw data (images) (Figure~\ref{fig:obj_representation}a). The HV in the item memories were sampled at the moment of model initialization and fixed during training. We used the same decoder architectures as in the other experiments.

\begin{figure*}[t]
    \centering
    \includegraphics[width=\textwidth]{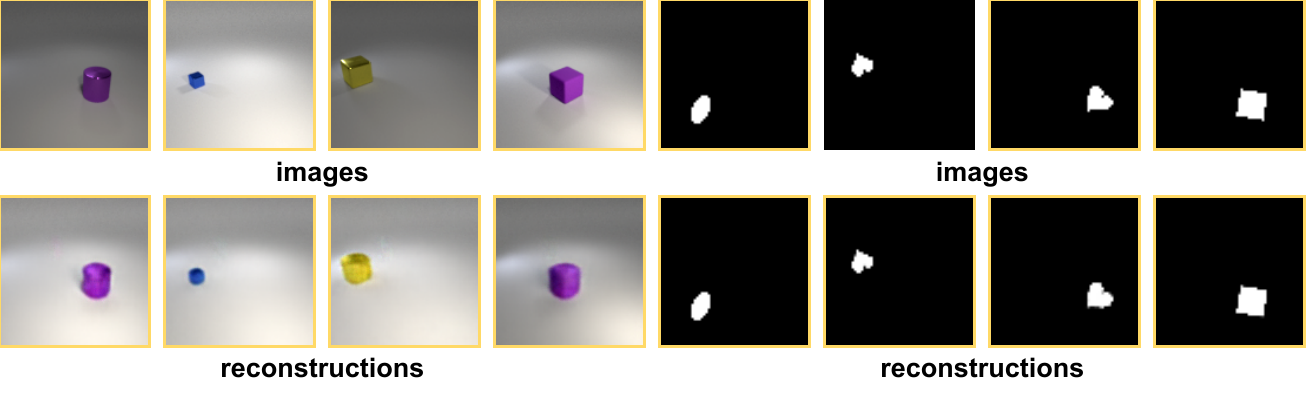}
    \caption{Image reconstruction of the CLEVR1 paired and dSprites paired objects from HDC representations without grounding to the images.}
    \label{fig:decoder_result}
    \vspace{10pt}
\end{figure*}

The results of the reconstruction for the HDC representation, obtained on the CLEVR1 paired and dSprites paired datasets without grounding to the images, are shown in Figure~\ref{fig:decoder_result}. For the CLEVR1 and dSprites datasets, the IOU and FID metrics (for 3 seeds) are 0.994$\pm$0.001 and 80.46$\pm$1.51, respectively.

\paragraph{RQ2. Is it possible to ground disentangled representations obtained with HDC to raw data using ArSyD?}
The answer to this question is the subject of the main set of experiments, in which we show how the proposed approach can be used to obtain disentangled representations based on HDC.

\begin{figure*}[t]
    \centering
    \includegraphics[width=\textwidth]{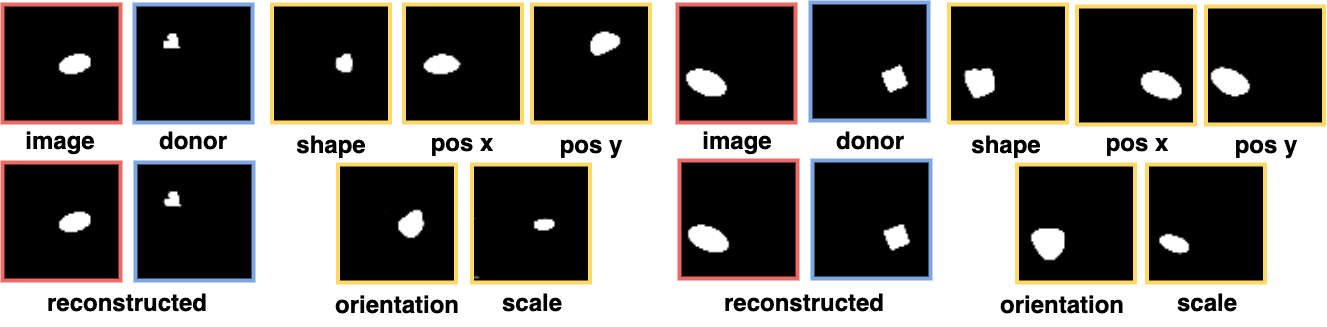}
    \caption{Image reconstruction of objects with modified generative factor values (yellow frame) for the \textbf{dSprites paired} dataset. The target object (red frame) differs from the donor object (blue frame) in all factor values.}
    \label{fig:dsprites_result}
    \vspace{10pt}
\end{figure*}

For the \textbf{dSprites paired} dataset, the results are shown in Table~\ref{tab:main_results} (bottom).
After training, we tested the possibility of exchanging features between random images (Figure~\ref{fig:dsprites_result}). The figure shows the results of the feature exchange for two training examples. On the left side of the image, the top left corner shows the two original images (ellipse and heart).  The second row below shows the result of their reconstruction. On the right, each image shows the result of the latent vector decoding after replacing a corresponding feature. It can be seen that the ``Shape'' feature is correctly replaced, but with a deformation. When ``Orientation'' and ``Scale'' are replaced, the main goal is achieved, but the shape of the object is slightly distorted in the case of ``Orientation''.

\begin{table}[ht]
    \centering
    \begin{tabular}[t]{c|r|r|r}
    \toprule
    \textbf{Metric} & \textbf{BetaVAE} & \textbf{FactorVAE} & \textbf{ArSyD (ours)} \\
    \midrule
    \multicolumn{4}{c}{\textbf{CLEVR1 paired}}\\
    \midrule
    FID $\downarrow$ & $129.68\pm13.21$ & $115.61\pm2.49$ & \textbf{93.72 $\pm$ 19.13} \\
    DMM $\downarrow$ & $0.99\pm00.12$  & \textbf{0.96 $\pm$ 0.27} & $1.10\pm00.04$ \\
    DCM $\downarrow$ & $2.74\pm00.33$ & $4.35\pm1.48$ & \textbf{1.88 $\pm$ 00.03} \\
    \midrule
    \multicolumn{4}{c}{\textbf{dSprites paired}}\\
    \midrule
    IOU(\%) $\uparrow$ & $96.43\pm0.29$ & $97.11\pm0.32$ & \textbf{98.37 $\pm$ 00.26} \\
    \bottomrule  
    \end{tabular}
    \vspace{5pt}
    \caption{For all models we used the same architectures for encoder and decoder. All results were computed for 3 seeds. The models were trained for 600 epochs on CLEVR1 paired and 200 epochs on dSprites paired.}
    \label{tab:main_results}
    \vspace{5pt}
\end{table}

On the right side of the image, the ellipse features are replaced one by one by the square features. Here we can see that the ``Orientation'' feature is strongly related to the ``Shape'' feature, unlike the ``Scale'', ``Pos X'', and ``Pos Y'' features, with which the images are restored relatively well. We explain this by the symmetry features of the figures: the square, the oval, and the heart are symmetric when rotated by $\pi/2$, $\pi$, and $2\pi$, respectively, while the possible rotation angles are in $[0, 2\pi]$.

In addition, the lower left corner of the image shows that the image is correctly reconstructed with the generative combination excluded from the training sample (square with coordinate $x > 0.5$). In both examples, the X and Y coordinates are correctly transferred from the donor image.

\begin{figure*}[ht]
    \centering
    \includegraphics[width=\textwidth]{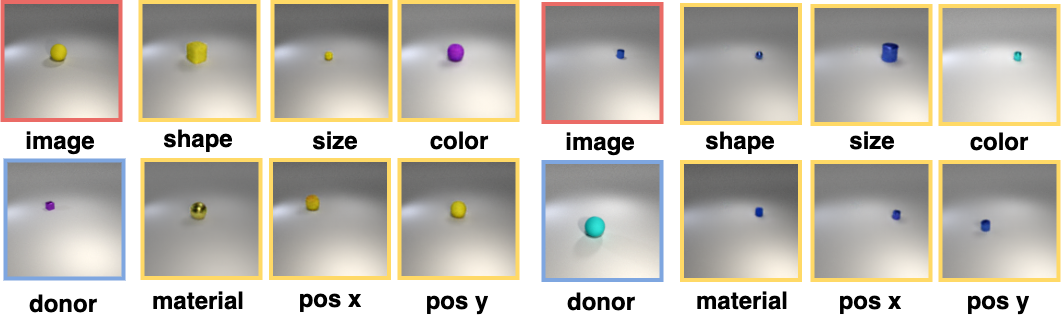}
    \caption{Image reconstruction of objects with modified generative factor values (yellow frame) for the CLEVR1 paired dataset. The target object (red frame) differs from the donor object (blue frame) in all factor values.}
    \label{fig:clevr1_result}
    \vspace{10pt}
\end{figure*}

\begin{figure*}[h]
    \centering
    \includegraphics[width=\textwidth]{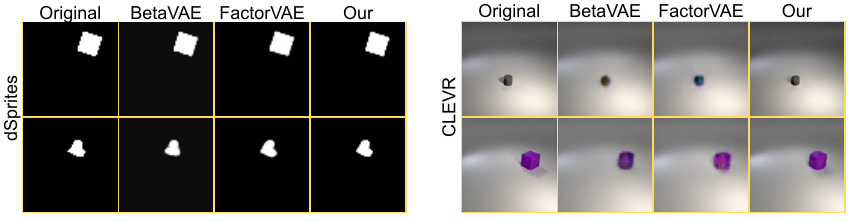}
    \caption{Examples of image reconstruction for all models from the dSprites paired (left) and CLEVR paired (right) datasets.}
    \label{fig:rec_beta_factor_compare}   
    \vspace{10pt}
\end{figure*}

The results for the \textbf{CLEVR1 paired} dataset, are shown in Table~\ref{tab:main_results} (top). The model has achieved a FID metric \cite{NIPS2017_8a1d6947} value of 103.16$\pm$2.76. Figure~\ref{fig:clevr1_result} shows the process of exchanging features between objects from the \textbf{CLEVR1 paired} dataset.  The feature exchange between random images works well with the ``Color'', ``Size'', and ``Material'' properties. In contrast to the \textbf{dSprites paired} dataset, there are problems with reconstruction when changing coordinates. The ``Shape'' feature exchange works well with respect to reconstructed images. We compare the reconstruction quality of ArSyD and baseline models in Figure~\ref{fig:rec_beta_factor_compare}.

\begin{figure*}[ht]
    \centering
    \includegraphics[width=\textwidth]{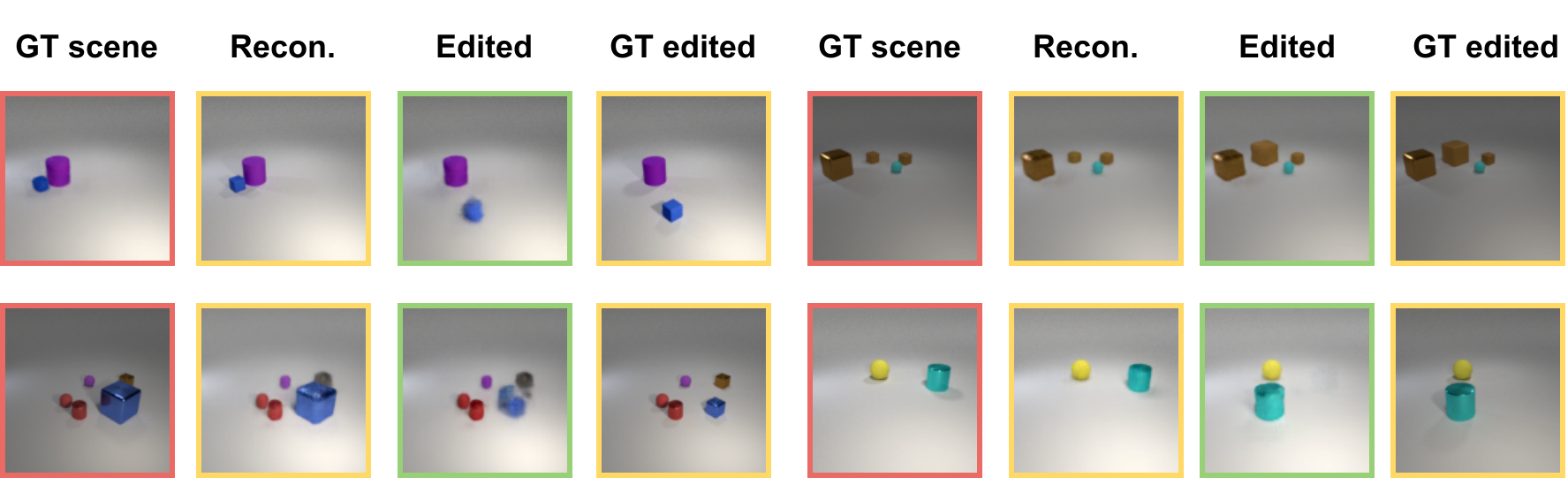}
    \caption{Image reconstruction of scenes with multiple objects from a test sample with modified values of generative factors of one object for the CLEVR5 paired dataset.}
    \vspace{10pt}
    \label{fig:clevr5_result}  
\end{figure*}

\paragraph{RQ3. Could ArSyD be used to modify a single object in a scene containing multiple objects?}
To answer this question, we used a combination of our approach and Slot Attention trained on the CLEVR5 paired dataset. The qualitative results for the \textbf{CLEVR5 paired} dataset are shown in Figure~\ref{fig:clevr5_result}.
The feature exchange between a single object in the scene generally works, except for the slot mask. Often, the restored slot mask for a modified image takes up more space than the restored object, which leads to problems during reconstruction: if there is a coordinate change, and one object is in front of the other one, the enlarged mask covers the first object.  In general, the quality of the reconstructed object from the slot is better than the decoding in CLEVR1, which is confirmed by a better FID metric equal to 67.19$\pm$1.6. The combination of ArSyD with Slot Attention allows object modification in a scene with multiple objects and correct reconstruction of scene images.

\paragraph{RQ4. How does the initialization of HV in the item memory affect overall performance?}
Since ArSyD uses randomly sampled HV that do not change during training, we tested the stability of the model for the dSprites dataset on $5$ seeds with HV dimension $D$ is equal to 1024. We obtained a mean of IoU is equal to $0.981$ with a variance of $0.002$. These results show that the specific set of vectors sampled during the initialization of ArSyD has little effect on the performance of the model.

\paragraph{RQ5. How does changing the HV dimension D affect overall performance?}
An important hyperparameter of the model is the dimension $D$ of the latent space. We tested how changing this value affects the performance of ArSyD for the dSprites paired dataset (Table~\ref{tab:ablation_dimensions}). It can be seen that, in general, the larger the dimension $D$, the better the performance. In this case, there is a trade-off between the computational complexity and the performance. As the dimension increases above 512, the metric values change insignificantly, therefore in our work we used the dimension $D=1024$ for the main experiments.

\begin{table}[ht]
        \centering
        \begin{tabular}[h]{c|c|c|c}
            \toprule
            \textbf{$D$} & \textbf{DMM}$\downarrow$ & \textbf{DCM}$\downarrow$ & \textbf{IoU} $\uparrow$\\
            \midrule
            16 & $1.46\pm0.12$& $1.74\pm0.02$& $0.948\pm0.025$  \\
            32 & $1.36\pm0.15$& $1.77\pm0.12$& $0.956\pm0.018$ \\
            64 & $1.33\pm0.13$&  $1.65\pm0.04$& $0.957\pm0.020$ \\
            128 & $1.24\pm0.02$& $1.71\pm0.07$& $0.968\pm0.001$ \\
            512 & \textbf{0.96 $\pm$ 0.07}& 1.58 $\pm$ 0.02& $0.973\pm0.004$\\
            1024 & 1.01 $\pm$ 0.04& 1.49 $\pm$ 0.08 & $0.984\pm0.001$\\
            2048 & 1.03 $\pm$ 0.06& \textbf{1.49 $\pm$ 0.03}& \textbf{0.986 $\pm$ 0.002} \\
            \bottomrule
        \end{tabular}
        \vspace{5pt}
        \caption{Metrics for the dSprites paired dataset with different latent dimension $D$. All results are computed for 3 seeds.}
        \label{tab:ablation_dimensions}
        \vspace{5pt}
\end{table}

\paragraph{RQ6. Is it possible to replace multiple generative factors at once with our approach?}
To answer this question, we performed an experiment with {1-4} simultaneous exchanges and obtained an IoU metric equal to 0.985 $\pm$ 0.004.

\begin{figure}[h]
    \centering
    \includegraphics[width=\linewidth]{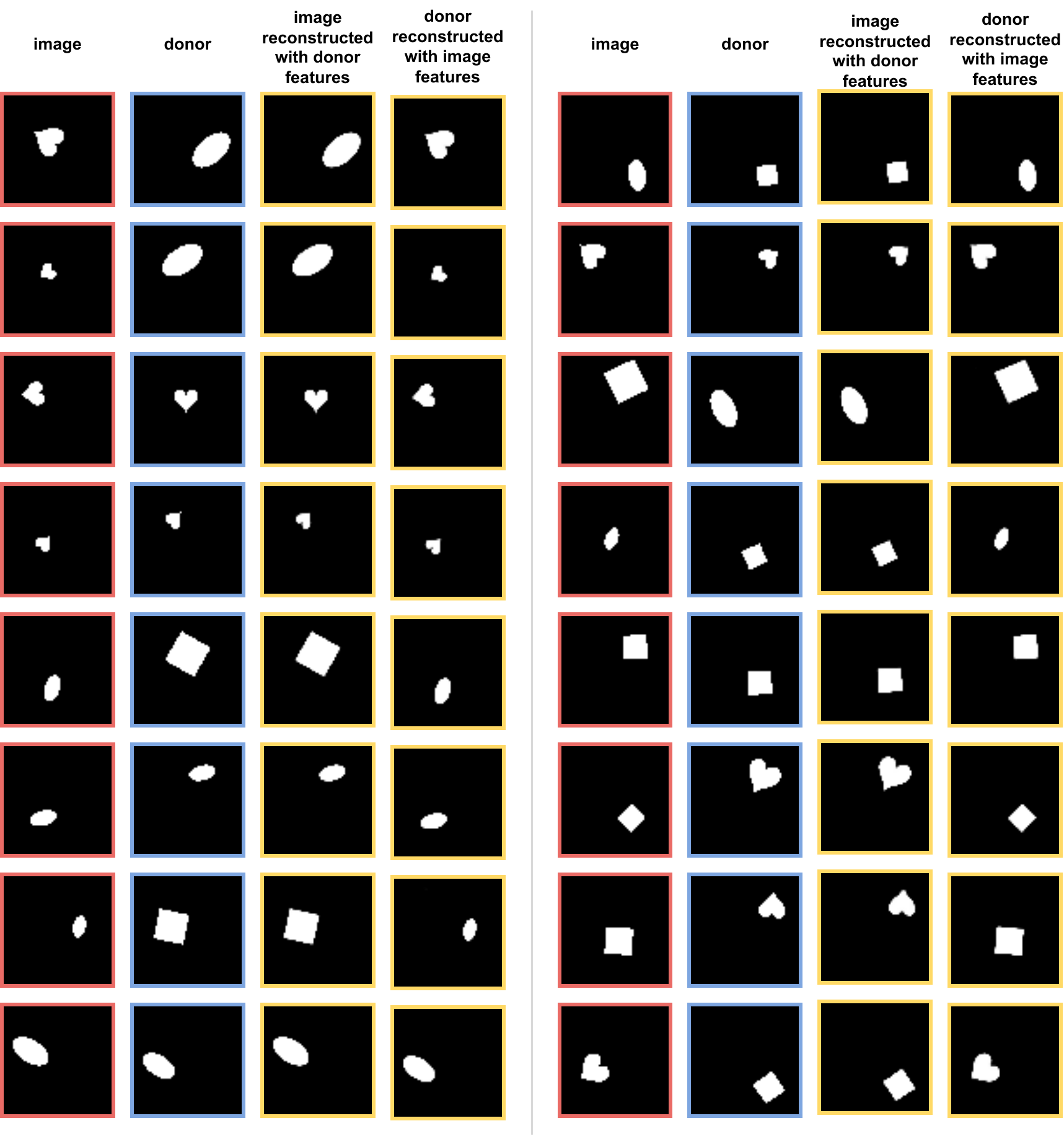}
    \caption{Examples of image reconstruction of objects from the dSprites paired dataset with modified values of the generative factors. The target object (red frame) differs from the donor object (blue frame) in all factor values.}
    \label{fig:multiple_exchanges_dsprites}
    \vspace{20pt}
\end{figure}

In Figure~\ref{fig:multiple_exchanges_dsprites}, we also show the qualitative results of the feature exchange for the case where the target object and the donor objects differ in all features. This demonstrates the generalizability of the proposed model for the case where only one feature differs.

\begin{figure}[h]
    \centering
    \includegraphics[width=\linewidth]{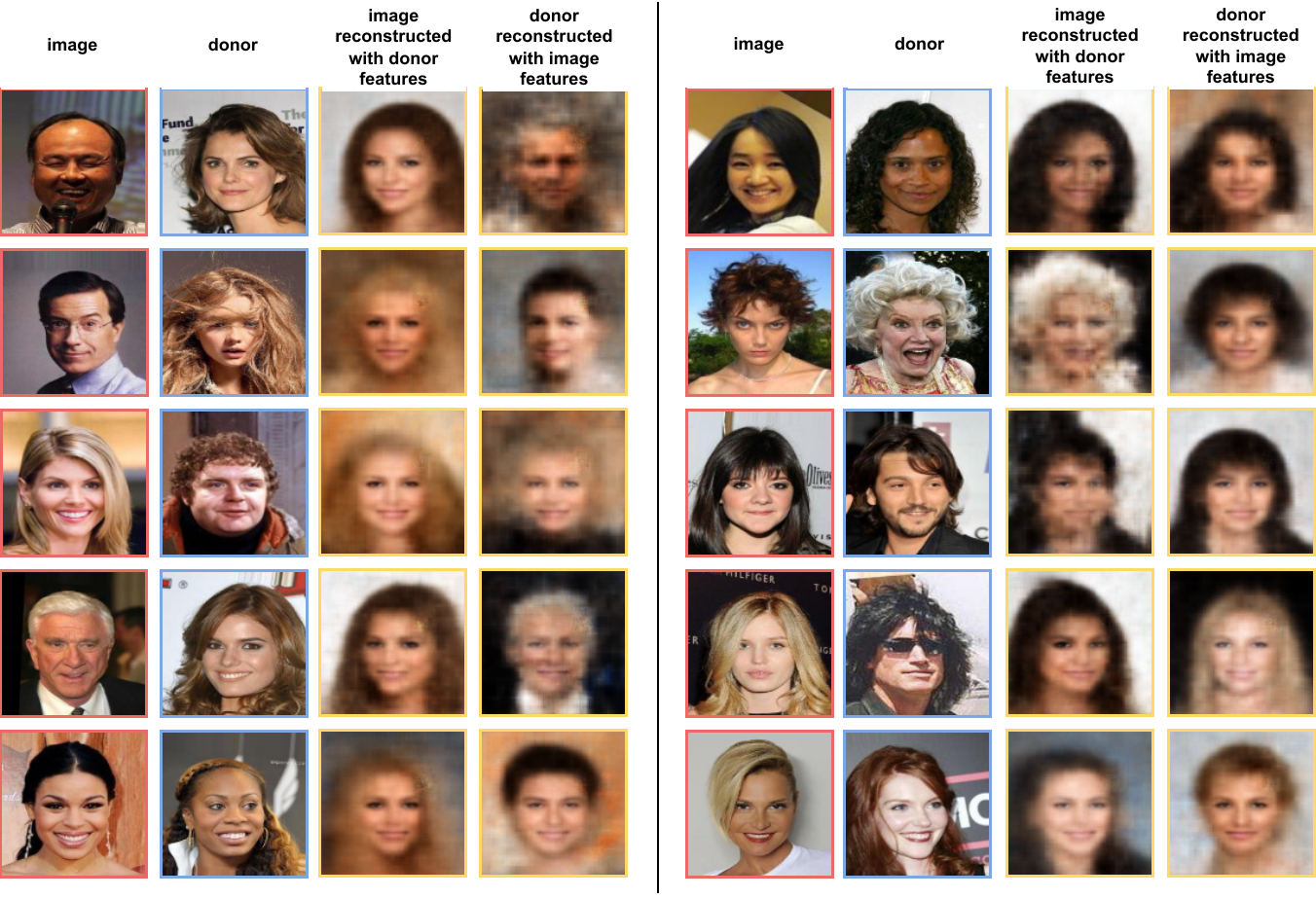}
    \caption{Examples of image reconstruction from the CelebA paired dataset.}
    \label{fig:celeba}
    \vspace{15pt}
\end{figure}

\paragraph{RQ7. Could the proposed approach be generalized to real-world data?}
To answer this question and to demonstrate the potential generalizability of ArSyD to real-world data, we conducted preliminary experiments on the CelebA dataset. For the experiments, we used the same architecture as for the CLEVR1 dataset. The reconstruction results are shown in Figure~\ref{fig:celeba}.

\begin{figure}[h]
    \centering
    \includegraphics[width=\linewidth]{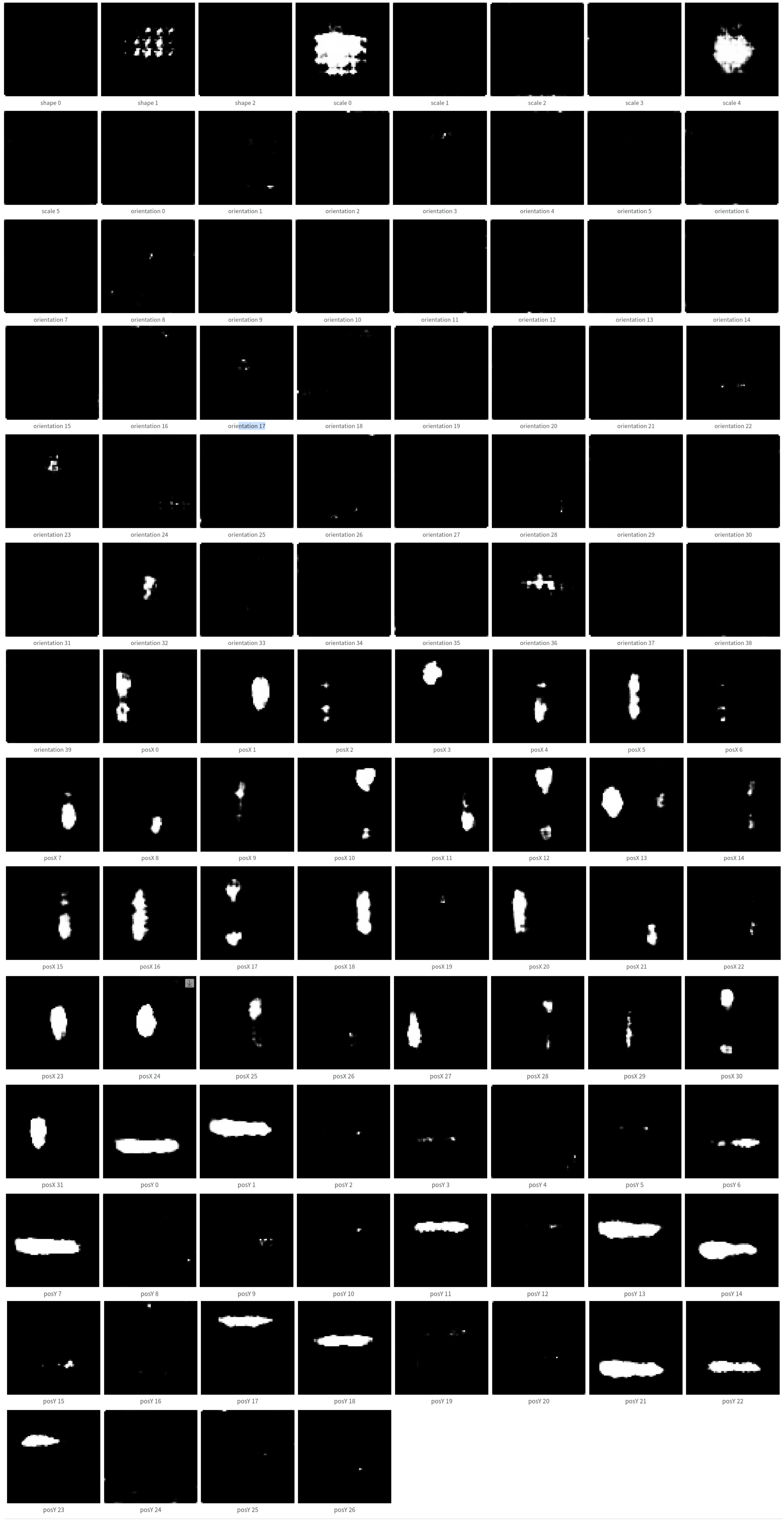}
    \caption{Visualization of image reconstruction from a single feature vector from an item memory (bound to a placeholder value) for the dSprites dataset. It shows that a complete image is not reconstructed from a single vector. This indicates that the vector represents a separate property of the object. This is particularly evident in the reconstruction of the position vectors. For example, on the bottom lines where Pos~Y is reconstructed, the line with the fixed position~Y is reconstructed.}
    \label{fig:vis_dsprites}
    \vspace{15pt}
\end{figure}

\begin{figure}[h]
    \centering
    \includegraphics[width=\linewidth]{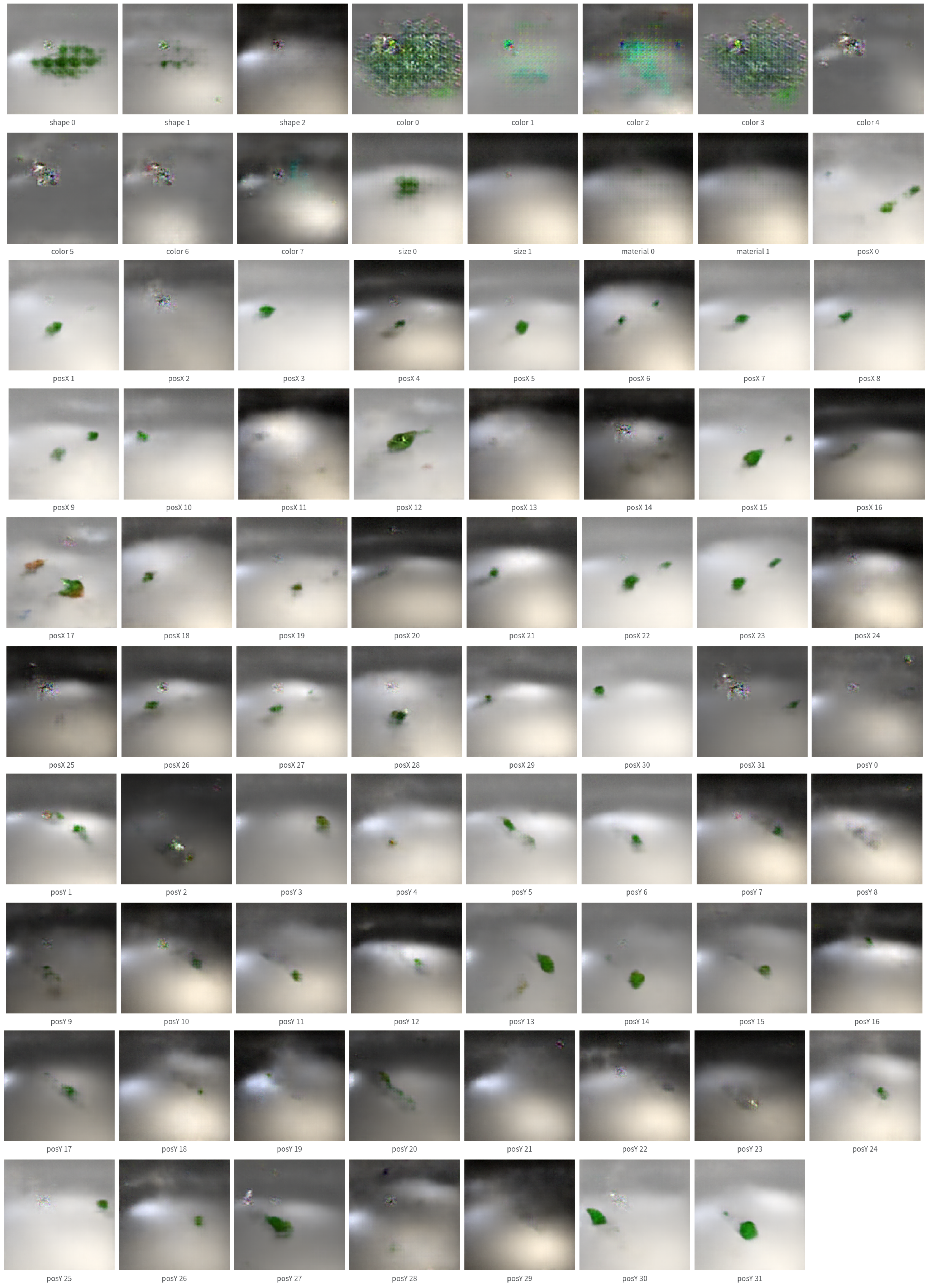}
    \caption{Visualization of image reconstruction from a single feature vector from an item memory (bound to a placeholder value) for the CLEVR dataset. It shows that a complete image is not reconstructed from a single vector. This indicates that the vector represents a separate property of the object.}
    \label{fig:vis_clevr}
    \vspace{15pt}
\end{figure}

\paragraph{RQ8. What information is contained in the resulting representations of the values of the generative factors?}
To answer this question, we reconstructed images from  a single feature vector from an item memory (bound to a placeholder value) for the dSprites paired and CLEVR paired datasets. Reconstruction results are shown in Figures \ref{fig:vis_dsprites} and \ref{fig:vis_clevr}. The results show that the vectors of the the values of the generative factors do not contain complete information about the object. Of particular interest are the reconstructed images for the generative factors "Pos X" and "Pos Y", where there is an approximate area on the image in which the object can appear for a given value of that generative factor.

\paragraph{RQ9. How does changing the HV dimension D affect the performance of classifiers used to compute DMM and DCM metrics?}
Since the results of the classification by the latent representation of the object are used to compute the proposed metrics, the size of this representation can affect the quality of the classification and thus the accuracy of the metric computation. Therefore, we conducted experiments with varying the dimension of $D$ and computed the classification accuracy for dSprites paired and CLEVR1 paired datasets. The results are shown in Tables~\ref{tab:acc_dsprites} and \ref{tab:acc_clevr}. The results show that in general, the higher the dimensionality $D$, the higher the classification accuracy. However, for spatial generative factors (orientation and position) with a large number of values, the classification quality is quite low. A possible solution could be to use a stronger classification model.

\begin{table}[h]
\fontsize{7pt}{8pt}\selectfont
        \centering
        \begin{tabular}[t]{ c|c|c|c|c|c}
            \toprule
            $D$ & \textbf{Shape} & \textbf{Scale} & \textbf{Orientation} & \textbf{Pos X} & \textbf{Pos Y} \\
            \midrule
16 & 0.36$\pm$0.08 & 0.23$\pm$0.31 & 0.09$\pm$0.06 & 0.04$\pm$0.03 & 0.01$\pm$0.01 \\
32 & 0.61$\pm$0.01 & 0.18$\pm$0.18 & 0.11$\pm$0.02 & 0.03$\pm$0.02 & 0.03$\pm$0.02 \\
64 & 0.41$\pm$0.12 & 0.32$\pm$0.12 & 0.17$\pm$0.06 & 0.15$\pm$0.04 & 0.05$\pm$0.02 \\
128 & 0.30$\pm$0.09 & 0.47$\pm$0.10 & 0.36$\pm$0.02 & 0.14$\pm$0.02 & 0.17$\pm$0.08 \\
256 & 0.24$\pm$0.19 & 0.53$\pm$0.11 & 0.37$\pm$0.04 & 0.20$\pm$0.04 & 0.09$\pm$0.03 \\
512 & 0.22$\pm$0.09 & 0.52$\pm$0.16 & 0.36$\pm$0.08 & 0.18$\pm$0.06 & 0.19$\pm$0.06 \\
1024 & 0.52$\pm$0.15 & \textbf{0.70$\pm$0.07} & 0.26$\pm$0.05 & 0.33$\pm$0.14 & 0.29$\pm$0.06 \\
2048 & \textbf{0.69$\pm$0.04} & 0.60$\pm$0.18 & \textbf{0.45$\pm$0.02} & \textbf{0.37$\pm$0.08} & \textbf{0.39$\pm$0.07} \\

            \bottomrule
        \end{tabular}
        \vspace{5pt}
        \caption{The classification accuracy for different latent space sizes $D$ for each of the generative factors for the dSprites paired dataset.}
        \label{tab:acc_dsprites}
        \vspace{10pt}
\end{table}

\begin{table*}[ht]
\fontsize{9pt}{10pt}\selectfont
        \centering
        \begin{tabular}[t]{ c|c|c|c|c|c|c}
            \toprule
            $D$ & \textbf{Shape} & \textbf{Size} & \textbf{Material} & \textbf{Color} & \textbf{Pos X} & \textbf{Pos Y} \\
            \midrule

        16 & 0.52 $\pm$ 0.33 & 0.51 $\pm$ 0.06 & 0.42 $\pm$ 0.41 & 0.27 $\pm$ 0.26 & 0.04 $\pm$ 0.02 & 0.04 $\pm$ 0.05 \\
        32 & 0.44 $\pm$ 0.08 & 0.58 $\pm$ 0.13 & 0.55 $\pm$ 0.14 & 0.19 $\pm$ 0.08 & 0.06 $\pm$ 0.02 & 0.09 $\pm$ 0.07 \\
        64 & 0.58 $\pm$ 0.11 & 0.73 $\pm$ 0.18 & 0.65 $\pm$ 0.24 & 0.18 $\pm$ 0.01 & 0.08 $\pm$ 0.03 & 0.10 $\pm$ 0.06 \\
        128 & 0.54 $\pm$ 0.09 & 0.91 $\pm$ 0.03 & 0.59 $\pm$ 0.04 & 0.45 $\pm$ 0.09 & 0.17 $\pm$ 0.05 & 0.25 $\pm$ 0.09 \\
        256 & 0.63 $\pm$ 0.06 & 0.94 $\pm$ 0.04 & 0.71 $\pm$ 0.06 & 0.43 $\pm$ 0.06 & 0.20 $\pm$ 0.10 & 0.33 $\pm$ 0.06 \\
        512 & \textbf{0.79 $\pm$ 0.07} & 0.96 $\pm$ 0.02 & 0.80 $\pm$ 0.04 & 0.52 $\pm$ 0.05 & 0.28 $\pm$ 0.09 & 0.39 $\pm$ 0.12 \\
        1024 & 0.75 $\pm$ 0.04 & \textbf{0.98 $\pm$ 0.01} & 0.89 $\pm$ 0.07 & 0.64 $\pm$ 0.03 & 0.37 $\pm$ 0.10 & \textbf{0.60 $\pm$ 0.03}\\
        2048 & 0.78 $\pm$ 0.05 & 0.96 $\pm$ 0.01 & \textbf{0.95 $\pm$ 0.01} & \textbf{0.70 $\pm$ 0.02}& \textbf{0.65 $\pm$ 0.02} & 0.55 $\pm$ 0.10 \\
            \bottomrule
        \end{tabular}
        \vspace{5pt}
        \caption{The classification accuracy for different latent space sizes $D$ for each of the generative factors for the CLEVR1 paired dataset.}
        \label{tab:acc_clevr}
        \vspace{10pt}
\end{table*}

\section{Limitations}
To train ArSyD, it is necessary to assume the number of generative factors in the data, since this is a model hyperparameter. This is not a limitation when working with synthetic data, such as game and simulation environments, if the number of generative factors is known in advance. However, when working with realistic data, additional analysis of the data itself may be required to determine which level of the representation hierarchy is worth using. For example, you can describe a person's face in terms of hairstyles and hair colors, nose shapes, eye colors, and so on, or you can go down the representation hierarchy and describe the shape of a nose, for example, in terms of tangent tilt angles to the line of the nose. The limitations associated with preliminary data analysis are not unique to our approach, and are also inherent in other object-centric models, such as those that use a slot representation~\cite{locatello2020object}. The number of slots in such a model is chosen as the maximum possible number of objects in the scene, with or without a background. Although the generalization of the proposed model to the case of realistic data is an interesting direction for further research, we nevertheless demonstrate the proof-of-concept results on the CelebA dataset.

The proposed metrics have the following limitations. First, it is necessary to know the number and type of generative factors in advance. Second, it is important to have a labeled dataset on which the classifiers can be trained. This can be problematic for real-world data. Furthermore, since we are using classifiers, their quality and robustness contribute significantly to the metric computation. This is further enhanced by the fact that the training is done on the original data (to avoid bias towards one or the other model being tested) and the classification for the metric is done on the reconstructed images. Another factor is the use of the decoder and its effect on the quality of the reconstruction. A better decoder is more likely to result in better classification quality. This influence is partially eliminated by using the same decoder architecture for different models and by considering relative changes in the prediction rather than absolute changes (we do not compare with the ground truth answers). Despite all these limitations, the proposed metrics are a convenient tool for comparing models that use different latent representations.

\section{Discussion}
An important issue is the generalization of the ArSyD to real-world data. To do this effectively, two different situations must be considered: one where the set of generative factors is known, and another where it remains unknown. In scenarios where the generative factors for each object in the dataset are known, symbolized as $G^i = G(O_i)$ -- $G^i $ is a disentangled representation of an object $O_i$, we can collect a set of training examples to initially teach the model to recognize these generative factors. If there are $N$ generative factors and two objects $O_1$ and $O_2$ differ by $k$ generative factors, then it is necessary to simultaneously exchange $k$ corresponding generative factors in $G^1$ and $G^2$ to train the model. Due to the significant combinatorial increase in possible combinations, this approach does not require the creation of a large dataset with a large amount of initially labeled data.

In scenarios without using labeled data, it is possible to use pre-trained neural networks, which are good feature extractors and focus on similarity when exchanging generative factors. For example, in the object-centric domain, breakthrough results on real-world data have been achieved by~\cite{seitzer2022bridging}, which uses the training signal from the feature reconstruction of a pre-trained DINO model~\cite{caron2021emerging}.

We consider extending our approach to the unlabeled data, where the selection of generative factors to exchange during training is based on a criterion, as future work. This could be a neural network-based criterion trained on a labeled set using examples where we know the differences, or a threshold-based criterion selected from examples by a similarity function. We also want to note that in practical situations, not all possible generative factors are usually considered, but a certain reasonable subset. For example, the type of hairstyle may serve as a generative factor, but not the generative factor responsible for the location of individual curls. The number of possible generative factors in localist models is also limited by the dimension of the hidden representation, and some of the coordinates may be redundant in these representations. In our approach, it is also possible to sample an initially redundant number of seed vectors, store them in item memory, and, as written above, use one of the possible criteria, e.g., cosine similarity, for the exchange.

The key component that allows bridging the gap between the localist representation obtained at the output of the encoder and the distributed representation used in HDC is the attention mechanism. The idea of an attention mechanism can be interpreted in terms of HDC. In HDC, a multiset can be represented as a simple set of vectors by summation, given that the necessary vectors appear multiple times in the sum. This is equivalent to multiplying the vector in the sum by a coefficient equal to the number of entries of that vector, or if the sum is normalized to the frequency coefficient, which corresponds to the attention weights in Equation~\ref{eq:att}. The resulting sum will be closer to the vector that appears more often in the sum.

One of the motivations for developing models that work with HDC is that it uses distributed representations that are highly resistant to noise. This potentially allows HDC models to be implemented on a nanoscale hardware, a class of devices that operate at ultra-low voltages because they consist of unreliable, stochastic computational elements. These devices are highly parallel, fabricated at ultra-small scales, consume less power than conventional devices, and are very advantageous for use in wearable devices or mobile robots~\cite{Kleyko2021}. The use of localistic representations obtained by models based on VAE or GAN on such devices is impossible due to the fact that feature values are encoded at specific positions of the vector and adding noise to these positions can lead to a drastic changes in the properties of an encoded object.

\section{Conclusion}
\label{sec:conclusion}
In this article, we propose ArSyD -- Architecture for Symbolic Disentanglement -- which learns symbolic disentangled representations using a structured representation of an object in latent space through a weakly supervised learning procedure. This representation is based on the principles of Hyperdimensional Computing, and we represent each generative factor as a hypervector of the same dimension as the resulting representation. The object representation is then obtained as a superposition of the hypervectors responsible for the generative factors. In contrast to classical approaches based on VAE or GAN, the learned representations are distributed, i.e., individual coordinates are not interpretable (unlike localist representations). Potentially, this allows us to reduce changing the properties of an object to manipulating its latent representation using vector operations defined in Hyperdimensional Computing. We also propose a way to edit the scene by changing the properties of a single object. This is achieved by combining the Slot Attention model, which allows us to find individual objects in the scene, and the proposed model with symbolic disentangled representations. To compare models that use latent representations of different structures, we propose two metrics that allow us to compare models that use latent representations of different structures: the Disentanglement Modularity Metric and the Disentanglement Compactness Metric.  The Disentanglement Modularity Metric refers to the evaluation of whether each latent unit is responsible for encoding only one generative factor. The Disentanglement Compactness Metric measures whether each generative factor is encoded by a single latent unit. The metrics are not applied directly to latent representations, but to already reconstructed images. We have shown how ArSyD learns symbolic disentangled representations for objects from modified dSprites and CLEVR datasets, and provide the proof-of-concept results on the realistic CelebA dataset.
\bibliography{arsyd}

\end{document}